\pdfoutput=1
\pdfoutput=1
\pdfoutput=1
\pdfoutput=1
\pdfoutput=1
\documentclass[twoside,twocolumn]{article}

\usepackage{blindtext} % Package to generate dummy text throughout this template 

\usepackage[sc]{mathpazo} % Use the Palatino font
\usepackage[T1]{fontenc} % Use 8-bit encoding that has 256 glyphs
\usepackage{microtype} % Slightly tweak font spacing for aesthetics

\usepackage[english]{babel} % Language hyphenation and typographical rules

\usepackage[hmarginratio=1:1,top=32mm,left=40pt, columnsep=20pt]{geometry} % Document margins
\usepackage[hang, small,labelfont=bf,up,textfont=it,up]{caption} % Custom captions under/above floats in tables or figures
\usepackage{booktabs} % Horizontal rules in tables

\usepackage{lettrine} % The lettrine is the first enlarged letter at the beginning of the text

\usepackage{enumitem} % Customized lists
\setlist[itemize]{noitemsep} % Make itemize lists more compact

\usepackage{abstract} % Allows abstract customization
\usepackage{titlesec} % Allows customization of titles
\renewcommand\thesection{\Roman{section}} % Roman numerals for the sections
\renewcommand\thesubsection{\roman{subsection}} % roman numerals for subsections
\titleformat{\section}[block]{\large\scshape\centering}{\thesection.}{1em}{} % Change the look of the section titles
\titleformat{\subsection}[block]{\large}{\thesubsection.}{1em}{} % Change the look of the section titles

\usepackage{fancyhdr} % Headers and footers
\usepackage{titling} % Customizing the title section

\usepackage{hyperref} % For hyperlinks in the PDF

\usepackage{CJK}

\usepackage{graphicx}

\usepackage{subfigure}

\usepackage{amsmath,amssymb,amsfonts}

\usepackage{algorithmic}

\usepackage{makecell}

\pretitle{\begin{center}\Huge\bfseries} % Article title formatting
\posttitle{\end{center}} % Article title closing formatting
\title{Improving Chinese Segmentation-free Word Embedding With Unsupervised Association Measure} % Article title
\author{%
\textsc{ZHANG Yifan\textsuperscript{1,2,3}},
\textsc{WANG Maohua\textsuperscript{2,3,4}}, 
\textsc{HUANG Yongjian\textsuperscript{2,3}},
\textsc{GU Qianrong\textsuperscript{2,3}}\\[1ex] % Your name
\normalsize 1.~University of Chinese Academy of Sciences, Beijing 100049, China \\ % Your institution
\normalsize 2.~Key laboratory of Low carbon conversion science and engineering, \\
\normalsize Chinese Academy of Sciences, Shanghai 201210, China \\
\normalsize 3.~Shanghai Carbon Data Research Center, Shanghai Advanced Research Institute, \\
\normalsize Chinese Academy of Sciences, Shanghai 201210, China \\
\normalsize 4.~Centre for Excellence in Brain Science and Intelligence Technology, \\
\normalsize Chinese Academy of Sciences, Shanghai 200031, China\\
\normalsize \href{mailto:zhangyifan2018@sari.ac.cn}{zhangyifan2018@sari.ac.cn} % Your email address
}
\date{\today} % Leave empty to omit a date
\renewcommand{\maketitlehookd}{%
\begin{abstract}
\noindent Recent work on segmentation-free word embedding(sembei) developed a new pipeline of word embedding for unsegmentated language while avoiding segmentation as a preprocessing step. However, too many noisy n-grams existing in the embedding vocabulary that do not have strong association strength between characters would limit the quality of learned word embedding. To deal with this problem, a new version of segmentation-free word embedding model is proposed by collecting n-grams vocabulary via a novel unsupervised association measure called pointwise association with times information(PATI). Comparing with the commonly used n-gram filtering method like frequency used in sembei and pointwise mutual information(PMI), the proposed method leverages more latent information from the corpus and thus is able to collect more valid n-grams that have stronger cohesion as embedding targets in unsegmented language data, such as Chinese texts. Further experiments on Chinese SNS data show that the proposed model improves performance of word embedding in downstream tasks.
\end{abstract}
}

\begin{document}

% Print the title
\maketitle

%----------------------------------------------------------------------------------------
%	ARTICLE CONTENTS
%----------------------------------------------------------------------------------------

\section{Introduction}
\begin{CJK}{UTF8}{gbsn}
\lettrine[nindent=0em,lines=3]{A} s a fundamental step for natural language processing(NLP), word embeddings is essential to many downstream tasks, such as part-of-speech tagging$^{[1, 2]}$ named entity recognition$^{[3]}$, and machine translation$^{[4]}$. Most existing word embedding models aim to learn dense vector of words that are segmented from the corpus$^{[5 - 8]}$. However, it is not always easy to extract words from unsegmented languages. For instance, unlike English, Chinese words are not naturally delimited by spaces or symbols. Therefore, words segmentation is a necessary preprocessing step for conventional word embeddings model in Chinese.

Recent Chinese word segmentation neural mo-dels$^{[9, 10]}$ usually require dictionaries or manually annotated resources, which are not easily obtained and timely updated, especially in the specific domain. Moreover, the performance of segmenters is still far from perfect$^{[11]}$, such as Out-Of-Vocabulary(OOV) problems and unsatisfying results in informal data. So the quality of learned Chinese word embeddings can be degraded by the previous segmented errors.

To deal with the problems mentioned above, segmentation-free word embedding model$^{[12]}$ called sembei which avoids segmentation in the preprocessing step has been proposed. However, occurrence frequency as the only n-gram selection criteria used in sembei seems to be very skewed and not discriminative$^{[13]}$.

Because raw frequency ignores the fact that inner connection within n-gram which is referred to as association strength is also essential to evaluate the validness of a n-gram to be a word. For instance, given an frequent Chinese p-hrase "苹果和香蕉" in corpora, sembei simply counts its frequency and collect all possible n-gram into the embedding vocabulary. Yet some n-grams like "果和" and "和香" are usually have very weak association strength and could crowd out other valid n-grams in the embedding vocabulary with a fixed size. This problem would limit the performance of trained word vectors. One way trying to handle this problem is using a naive word boundary predictor to help collect more word-like n-grams$^{[14]}$. But this method is still dependent on the results segmentated by a supervised segmenter which is not consistent with the segmentation-free principle. Another work trains sub-n-gram level vectors and use them to represent the frequent words$^{[15]}$ in the vocabulary, but the same problem as sembei still remains. To our knowledge, few studies have yeild on improving segmentation-free word embedding with unsupervised method by increasing valid n-grams that have stronger association strength in the vocabulary.

In this paper, a new segmentation-free word em-bedding model called PATI Filtered N-grams Embedding(PFNE) is proposed. Specifically, a unsupervised n-gram association measure called pointwise association with times information(PATI) is proposed to replace frequency in sembei. Since no external resource is used to compute PATI, PFNE incorporates the advantage of unsupervised association measure into segmentation-free word embedding model. It is especially effective in unsegmented language rich in neologisms like Chinese SNS texts. Experiments on Chinese SNS data show that our method increases the number of valid n-grams in segmentation-free word embedding model by 12.6\% and 17.7\% comparing with sembei based on two dictionaries and improves F-1 score by 7.3\%, 3.7\% and 1.9\% in downstream tasks comparing with several baseline systems.

The following passage are organized as follows. In Section \uppercase\expandafter{\romannumeral2} the related works about segmentation-free word embeddings are introduced. And then, we present the new unsupervised assciation measure PATI and segmentation-free word embedding model PFNE in Section \uppercase\expandafter{\romannumeral3}. The experiments setting are described in Section \uppercase\expandafter{\romannumeral4} and the experiment results are shown in Section \uppercase\expandafter{\romannumeral5}. Finally, the conclusions and future work are presented in Section \uppercase\expandafter{\romannumeral6}.
\end{CJK}
%------------------------------------------------

\section{RELATED WORK}

Recently, models on representation of words that do not need any dictionaries or manually annotated resource have been proposed. Some models aim to solve task-specific problems, such as word segmentation$^{[16]}$, machine translation$^{[17]}$, new words detection$^{[18]}$ and texts classification$^{[19]}$. Some models aim to deal with domain-specific problems, such as representation of biological sequences$^{[20]}$, social media opinion mining$^{[21]}$ and clinical text correction$^{[22]}$. As for intrinsic word embedding problems, some models focusing on subword level embedding, such as Subword Information Skip Gram$^{[23]}$, Bag of subword model$^{[24]}$ and CHARANGRM model$^{[25]}$. But all these word embedding models are not practicable in the unsegmented languages because they all require segmented word as preprocessing step. 

Considering the situation of unsegmented language like Chinese, a new pipeline for word embedding model without segmentation$^{[26]}$ called sembei was proposed. It selects top-K frequent n-grams instead of segmented words as word embedding vocabulary. And then the co-occurrence information of these n-grams is used to train the embedding model. Experiments results proved its outperformance comparing with conventional skip-gram model with negative sampling on downstream tasks. But too many invalid n-gram existing in the vo-cablary brings down the performance of sembei in the downstream tasks. Identical problem also lies in the model which uses all sub-n-grams vectors to represent words, sentences and texts$^{[27]}$. One way to deal with this problem is performed in Ref.[28]. In this model, raw frequency is replaced by a metric called expected word frequency(ewf)$^{[29]}$ which is computed by a simple supervised word boundary probabilistic perdictor. PMI meausre is used as a explantory variable in training the predictor. Although it improves the word coverage in the vocabulary, word segmented by external segmenters and n-gram evaluation metric computed by supervised predictors are still required as necessary steps. 

\section{THE PROPOSED METHOD}
\subsection{THE RESEARCH FRAMEWORK}
In order to solve this problem by collecting more strongly associated n-grams with unsupervised method, we proposed a new segmentation-free word embedding model called PATI Filtered N-gram Embedding (PFNE) where PATI is the new  criteria proposed to evaluate association strength between characters in the n-grams. As shown in Figure 1, in PFNE model, all possible n-grams were collected as candidate n-gram for PATI evaluation and then top-K n-grams with the highest PATI score were selected as embedding vocabulary and used to construct n-grams lattice that contains word-context pairs. Subsequently, co-occurrence information over the n-gram lattice was fed into the embedding model to learn n-gram embeddings.
\begin{figure}
	\centering
	\includegraphics[]{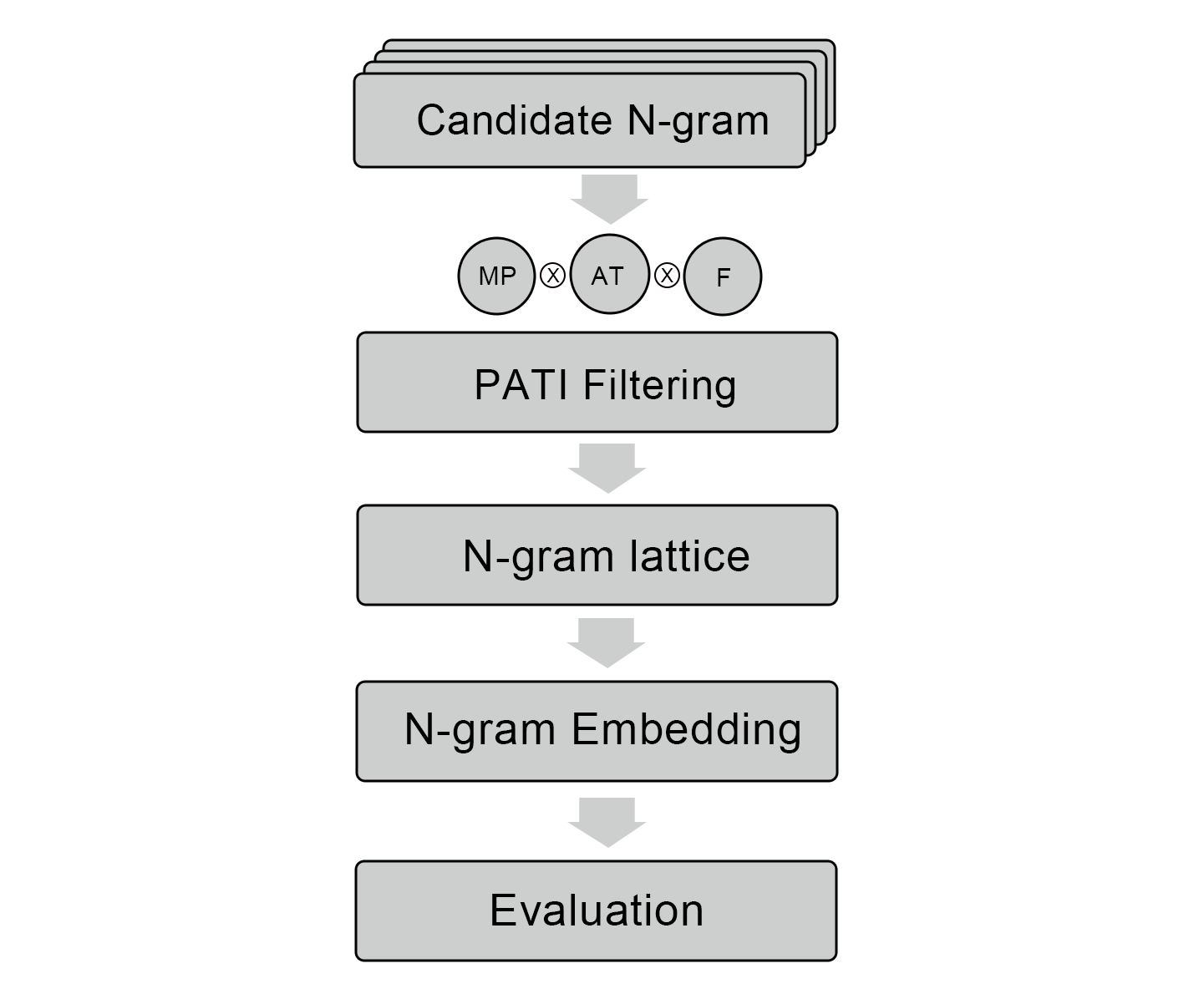}
	\caption{Fig.\ 1.\ The overview of PFNE.}
\end{figure}

\subsection{POINTWISE ASSOCIATION WITH TIMES INFORMATION}
As pointed out in Section \uppercase\expandafter{\romannumeral2}, solely using raw frequency as selection standard seems to be not very effective in filtering n-grams because it can not evaluate the association strength which is important to determine whether a n-gram is valid or not. Therefore, cohesion of n-grams should be considered to reduce the number of invalid n-grams as embedding targets. In order to balance the advantage of unsupervised method in handling the OOV problem and the purpose of avoiding words segmentation as a processing step in word embedding model, a new unsupervised association measure called pointwise association with times information(PATI) was proposed by considering more properties from the corpus.

In a Chinese corpus $C$ = $\left\lbrace w_{1}, w_{2}, w_{3} ,..., w_{N-1}, w_{N}\right\rbrace $ with size ${N}$,\, for a n-gram $g$ = $w_{i}w_{i+1}...w_{i+s}$ ($ 0 \le i \le N-s$) with fixed length $s$, $s$ usually ranges from 1 to 6. For $k\in(i, i+s)$, $g$ = $concat(a, b)$, n-gram segments $a = w_{i}...w_{k-1}$ and $b = w_{k}...w_{i+s}$ are all possible left and right part of the n-gram $g$. $f_{a}$, $f_{b}$, and $f_{g}$ are raw frequency of single n-gram segment $a$, $b$ and combined n-gram $g$ in the whole corpus. Then PATI is defined as follows:

\begin{equation}
PATI =  F \times MP \times AT  \label{eq}
\end{equation}

Here, $F$, $MP$ and $AT$ are three components in the formula. We will explain each component in the following steps.

{\bf Step 1: First component $F$}

$F$ is the raw frequency of n-gram $g$ = $concat(a, b)$ which is also used in sembei. N-grams with high frequency are more likely to be commmly used words in the corpus. $F$ is defined as:
\begin{equation}
F = f_{g}
\end{equation}

{\bf Step 2: Second Component $MP$}

For a n-gram $g$ = $concat(a, b)$ and all its possible left and right part $a$ and $b$, $MP$ value of n-gram $g$ is defined as:
\begin{equation}
MP = \min \left\lbrace \frac{(N \times f_{g})^2}{(f_{a}+f_{b})^2 \times f_{a} \times f_{b}} \right\rbrace 
\end{equation}

Given a n-gram $g$ = $concat(a, b)$ with fixed length $s$, there always exists a specific combination of n-gram segments $(a_{m}, b_{m})$ that minimize $MP$. Then the third component $AT$(Eq.(10)) in $PATI$ is computed under this combination $(a_{m}, b_{m})$.

{\bf Step 3: Third component $AT$}

Several concepts need to be introduced at first for computation of $AT$. \\For the specific combination $(a_{m}, b_{m})$ of each n-gram $g$, n-grams $(a_{m}, b_{h})$ and $(a_{j}, b_{m})$ have the same fixed length $s$ as n-gram $g$. Then two sets $\left\lbrace a_{m}, *\right\rbrace$ and $\left\lbrace *, b_{m}\right\rbrace$ are defined as:

\begin{equation}
\left\lbrace a_{m}, *\right\rbrace  = \left\lbrace (a_{m}, b_{1}), (a_{m}, b_{2}),..., (a_{m}, b_{h})\right\rbrace
\end{equation}

\begin{equation}
\left\lbrace *, b_{m}\right\rbrace  = \left\lbrace (a_{1}, b_{m}), (a_{2}, b_{m}),..., (a_{j}, b_{m})\right\rbrace 
\end{equation}

Let $f_{a_{m}*}$ and $f_{*b_{m}}$ represent the frequency of $\left\lbrace a_{m}, *\right\rbrace $ and $\left\lbrace *, b_{m}\right\rbrace$ which are as follows:

\begin{equation}
f_{a_{m}*} = \sum_{n=1}^{h} f_{a_{m}b_{n}}
\end{equation}

\begin{equation}
f_{*b_{m}} = \sum_{q=1}^{j} f_{a_{q}b_{m}}
\end{equation}

Let $rate$ be the maximum between the ratio of $f_{g}$ with $f_{a_{m}*}$ and ratio of $f_{g}$ with $f_{*b_{m}}$. Formula of $rate$ is defined as:

\begin{equation}
rate = \max\left\lbrace\frac{f_{g}}{f_{a_{m}*}}, \frac{f_{g}}{f_{*b_{m}}} \right\rbrace 
\end{equation}

Let $AC$ be the reciprocal of the size of set $\left\lbrace a_{m}, *\right\rbrace $ or $\left\lbrace *, b_{m}\right\rbrace $ which has the less frequency. Then $times$ and $AT$ are formulated as

\begin{equation}
times = \frac{rate}{AC}
\end{equation}

\begin{equation}
AT = 1+\vert\log{times} \vert
\end{equation}

Eventually, a math operation was did to combine $F$, $MP$ and $AT$ by multiplication.

\begin{CJK}{UTF8}{gbsn}
\begin{table}
	\centering
	\caption{\bf Table 1.\ Examples of valid 2-grams that are filtered out by MP while ignored by PMI (English name for reference).}
	\setlength{\tabcolsep}{3mm}
	\begin{tabular}{|c|c|c|}
		\hline
		2-gram & Name(en)      & PMI score \\ \hline
		银票     & Bandar’s note & 0.98      \\ \hline
		真知     & Truth         & 0.95      \\ \hline
		交情     & Fellowship    & 0.81      \\ \hline
		金主     & Investor      & 0.80      \\ \hline
		海产     & Seafood       & 0.78      \\ \hline
	\end{tabular}
\end{table}
\end{CJK}

$MP$(Eq.(3)) is the modified version of pointwise mutual information(PMI), where $((N^2)/(f_{a}\times f_{b}))$ is reciprocal of joint probability of n-gram $g$ in the corpus and $((f_{g})/(f_{a}+f_{b}))^2$ increases sensitivity to local information around the n-gram by considering marginal variables in association computation. Taking 2-gram as an example, as shown in Table 1, comparing with PMI, the modified version can find out many valid n-grams whose PMI scores are less than 1, which will not be seen as strongly associated n-grams by PMI in the view of statistics.

For a n-gram $g$ = $concat(a, b)$ and all its possible left and right part $a$ and $b$, $PMI$ value of n-gram $g$ is defined as:
\begin{equation}
PMI =\min \left\lbrace \frac{N \times f_{g}}{f_{a} \times f_{b}}\right\rbrace
\end{equation}

$AT$(Eq.(10)) is proposed to lervage statistic property between the specific n-gram combination $(a_{m}, b_{m})$ and set $\left\lbrace a_{m}, * \right\rbrace$ or set $\left\lbrace *, b_{m} \right\rbrace $. Variable $times$(Eq.(9)) in $AT$ indicates the relative importance of the n-gram $(a_{m}, b_{m})$ in set $\left\lbrace a_{m}, *\right\rbrace $ or $\left\lbrace *, b_{m}\right\rbrace $. The higher $times$ is, the more possible that $(a_{m}, b_{m})$ is a valid n-gram. And $times$ value of most valid n-grams should be much higher than those which are not valid. An example of computation of $times$ of valid and invalid 2-grams is shown in Fig.2.

\begin{figure}
	\centering
	\includegraphics[]{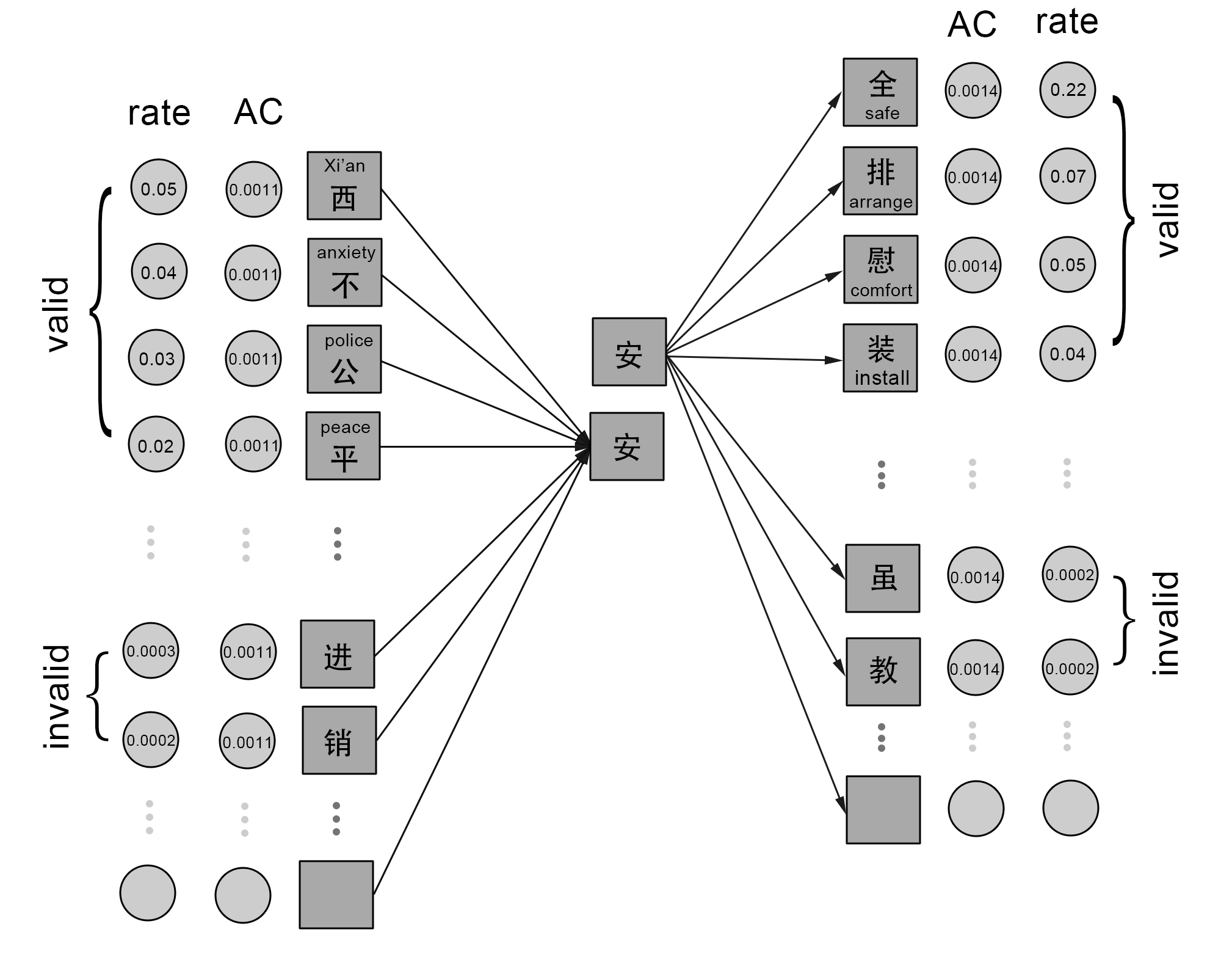}
	\caption{Fig.\ 2.\ Examples of computation of $times$ property. $times$ is ratio of $rate$ to $AC$. N-grams with higher $times$ means they are are more likely to be valid.}
\end{figure}

The new n-gram association measure PATI considers more latent statistic information for each candidate n-gram from the corpus. For example, as seen in Fig.3, the proposed method tends to include more valid n-grams than commonly used basic segmentation dictionary, sembei and PMI.

\begin{figure}[h]
	\centering
	\includegraphics[]{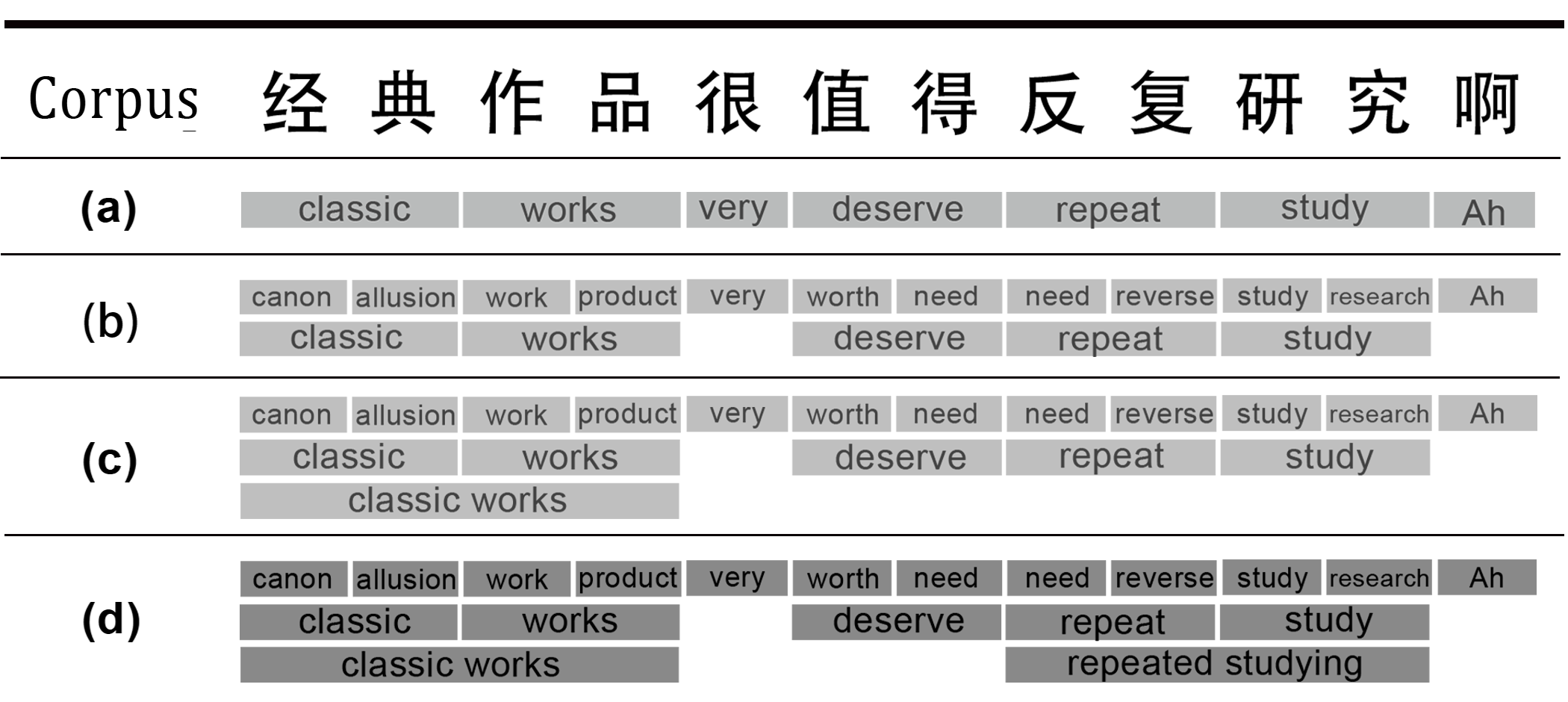}
	\caption{Fig.\ 3.\ A Chinese sentence in online community with manual segmentation. (a) is segmentation result of a basic segmentation dictionary which is used in conventional word embedding methods. (b), (c) and (d) are valid n-grams filtered by sembei, PMI and PATI.}
\end{figure}

\subsection{PATI FILTERED N-GRAM EMBEDDING MODEL}
After filtering candidate n-grams with PATI, a group of strongly associated n-grams was selected as embedding vocabulary. Then the n-gram lattice is constructed by collecting word-context n-gram pair form corpous according to embedding vocabulary. PFNE modified sembei with skip-gram negative sampling $^{[6]}$ by changing the definition of bag of positive samples $N_{p}$ from the pairs containg top-K frequent n-grams to pairs $(w_{t}, w_{c})$ that constructed by top-K n-grams that have the highest PATI scores. Correspondingly, the set of negative samples $N_{n}$ were also redefined by random sampling. The objective function of PFNE is:

\begin{equation}
\begin{aligned}
L\!=\!\sum_{(w_{t}, w_{c})\in N_{p}}\! \log(\!1+&\!e^{-\boldsymbol{x}_{w_{t}}^\top \boldsymbol{x}_{w_{c}}}) \!\\&+\! \sum_{(w_{t}, w_{c})\in N_{n}} \log(1+e^{\boldsymbol{x}_{w_{t}}^\top \boldsymbol{x}_{w_{c}}})
\end{aligned}
\end{equation}

where $\boldsymbol{x}_{w_{t}}$ and $\boldsymbol{x}_{w_{c}}$ are vector representation of word n-grams $w_{t}$ and context n-grams $w_{c}$. We optimize this function with stochastic gradient descent(SGD)$^{[26]}$.

\section{EXPERIMENT SETUP}

In this section, we would like to describe our experimental data, models and experimental setup. The implementation of proposed method is available on GitHub$^{**}$\footnote{$**$
	\emph{https://github.com/zyfIvan1997/PFNE.git}}.

\subsection{Data Sets}

There are mainly 2 data sets and 2 lexicon data used in experiments:
\begin{itemize}
	\item 100MB SNS data$^{[27]}$ about online commmunity contains 39,543,712 Chinese characters.
	\item Wikidata$^{[28]}$ (We used dumps\footnote{https://dumps.wikimedia.org/wikidatawiki/entities/20200224/} dated February, 24th, 2020.).
	\item Words dictionary named basic dictionary\footnote{https://github.com/fxsjy/jieba/blob/master/extra\_dict/dict.txt.small.} that contains 109750 words used in Jieba Segmenter.
	\item Words dictionary named rich dictionary\footnote{https://github.com/fxsjy/jieba/blob/master/extra\_dict/dict.txt.big.} that contains 584429 words used in Jieba Segmenter.
\end{itemize}

\subsection{Models}
Two main experiments are conducted in this paper. The first is about the n-gram seletcion of the segmentation-free word embedding model and the second is about the performance of word embedings in the downstream task. So, three segmentation-free methods: sembei, PMI and PFNE combined with Skip-gram model with Negative Sampling (SGNS) are evaluated in the first experiment. While in the second experiment, except for those three methods mentioned above, Skip-Gram in the word2vec which is a widely used word embedding model is also added as another strong baseline. 

For sgementation-free models, the dimension of word embedding is fixed to 200 and the number of iterations is 5. Initial learning rate $\alpha_{init}$ is set to 0.01, and the size of context window h is fixed to 1. We used number of negative samplings $\eta_{neg}$=10.

For Skip-Gram based on word segmenter, most settings are same as segmentation-free models. A grid search over $h \in \left\lbrace 1, 5, 10\right\rbrace$  is performed, where $h$ is the context window size.

{\bf (1) Baseline}
\begin{itemize}
	\item Skip-Gram: The most general word embedding model proposed in Ref.[6]. In this paper, Skip-gram with negative sampling is used based on a standard segmenter\footnote{jieba with jieba/dict.txt.small.}.
	
	\item SGNS-sembei: model proposed in Ref.[12]. In this model, the n-gram vocabulary is constructed by the top-K most frequent n-grams. Then, the occurrence information of n-grams lattice constructed by n-gram vocabulary is used to train word embedding.
	
	\item SGNS-PMI: In this model, the n-gram vocabulary is selected by PMI measure. The top-K n-grams with the highest PMI score is considered as embedding targets.
\end{itemize}

{\bf (2) Proposed model}

\begin{itemize}
	\item SGNS-PFNE:SGNS-PFNE is proposed by replacing n-gram raw frequency used in SGNS-sembei with PATI. Then top-K n-grams with the highest PATI score are selected as embedding vocabulary.
\end{itemize}

\subsection{Experiments}

The proposed method aims to improve the performance of word vectors of segmentation-free word embedding model for Chinese texts by increasing valid n-grams with strong association strength in the embedding vocabulary. Therefore, in this section, we conducted two major experiments. The first experiment is to examine the number of valid n-grams selected by those three segmentation-free word embedding methods. The second experiment is to verify the performance of word embeddings with noun category prediction task.

{\bf Experiment \uppercase\expandafter{\romannumeral1} : N-grams Selection Criteria}

N-gram raw occurrence frequency$^{[29]}$ and pointwise mutual information (PMI)$^{[30]}$ are two commonly used criteria to extract n-gram in computational linguistic. The principle of raw occurrence frequency used in sembei is just to count the times of n-grams that appear in the texts. Pointwise mutual information is a criteria based on information entropy. 

In this experiment, valid n-grams are defined as n-grams in basic or rich dictionary and invalid n-grams are defined as those not in the dictionaries. We examined the number of valid n-grams selected by sembei, PMI and PFNE. N-grams that occur less than 2 times were ignored. Exact numbers are listed from 2-gram to 6-gram based on $100$MB SNS data at a fixed vocabulary size $K$= 1,005K. In addition, Prescision-Recall curves was used to show the comparison by changing size of vocabulary from $1$ to $1300$K. The Precision and Recall are computed as follows:
\begin{equation}
Precision = \dfrac{Number \ of \ valid \ n-grams}{size\ of\ vocabulary}
\end{equation}
\begin{equation}
Recall = \dfrac{Number \ of \ valid \ n-grams}{size\ of\ dictionary}
\end{equation}

{\bf Experiment \uppercase\expandafter{\romannumeral2}: Noun Category Prediction}

Word vectors can capture semantic information from the corpus. The main role of word vectors is used as features in downstream NLP tasks. So noun category prediction task based on the trained word embeddings was performed to verify the improvement of effect of our method. Most of the settings are the same as Ref.[12]. Nouns with predetermined category\footnote{\{chemical compound, profession, taxon, city, country, company, human\}}  were extracted from Wikidata. N-grams in the nouns-category pair set were split into train $(60\%)$ and test $(40\%)$ sets. Then a linear C-SVM was trained to predict category of n-grams according to their learned embeddings. 1,005K n-grams\footnote{In this experiment, we define the embedding vocabulary as the union of top-$K_{n}$ n-grams with highest PATI score, the n is set to 6 and \{ $K_{1}$,$K_{2}$,$K_{3}$,$K_{4}$,$K_{5}$,$K_{6}$ \} are set to \{5000, 300000, 300000, 300000, 50000, 50000\} } was selected as the size of the embedding vocabulary. A grid search over $\left( C, classfier\right) \in \left\lbrace 0.5, 1.0, 1.5, 10, 50\right\rbrace \times \left\lbrace 1-vs.-1, 1-vs.-all \right\rbrace $ was conducted on the linear C-SVM. Weighted average Precision, Recall and F1 score were used as evaluation which are computed as follows:

\begin{equation}
Precision = \sum_{i=1}^{i=N}\alpha_{i}\dfrac{TP_{i}}{TP_{i}+FP_{i}}
\end{equation}
\begin{equation}
Recall = \sum_{i=1}^{i=N}\alpha_{i}\dfrac{TP_{i}}{TP_{i}+FN_{i}}
\end{equation}
\begin{equation}
F1 = \sum_{i=1}^{i=N}\alpha_{i}\dfrac{2\times Prescision_{i}\times Recall_{i}}{Precision_{i}+Recall_{i}}
\end{equation}
where $\alpha_{i}$ is the proportion of $i_{th}$ classification in all 
classification. $TP$ is the true positive, $FP$ is the false positive and $FN$ is the false negative.

\section{RESULTS}

\subsection{Results of experiments \uppercase\expandafter{\romannumeral1}}
The results of experiment 1 are shown in Table 2 and Figure 4. The exact number of valid n-gram in SGNS-sembei, SGNS-PMI and SGNS-PFNE and theirs percentage comparing with PFNE are listed repesctively in Table 2(a) and Table 2(b) by taking basic dictionary and rich dictionary as reference. PR-curves of these three methods with reference to basic dictionary and rich dictionary are also shown respectively in Figure 4(a) and Figure 4(b).

\begin{table}[!t]
	\caption{Results of n-grams selection criteria.}
	\begin{center}
		{\bf (a) result on the basic dictionary}
		\begin{minipage}[t]{0.95\textwidth}
			\begin{tabular}{|c|c|c|c|c|}
				\hline
				N-gram & size& sembei& PMI& PFNE  \\ \hline
				1-gram & 
				5,000& 
				\makecell[c]{5000\\(100)}& 
				\makecell[c]{5000\\(100)}&
				\textbf{\makecell[c]{5000\\(100)}} \\ \hline
				2-gram& 
				300,000& 
				\makecell[c]{36791\\(90.21)}&
				\makecell[c]{38551\\(94.53)}&
				\textbf{\makecell[c]{40783\\(100)}} \\ \hline
				3-gram& 
				300,000& 
				\makecell[c]{6460\\(70.72)}&
				\makecell[c]{7180\\(78.53)}&
				\textbf{\makecell[c]{9134\\(100)}} \\ \hline
				4-gram& 
				300,000& 
				\makecell[c]{4978\\(83.98)}&
				\makecell[c]{5657\\(95.44)}&
				\textbf{\makecell[c]{5927\\(100)}} \\ \hline
				5-gram& 
				50,000& 
				\makecell[c]{159\\(83.99)}&
				\makecell[c]{189\\(90.70)}&
				\textbf{\makecell[c]{217\\(100)}} \\ \hline
				6-gram& 
				50,000& 
				\makecell[c]{96\\(88.07)}&
				\textbf{\makecell[c]{109\\(105.83)}}&
				\makecell[c]{103\\(100)} \\ \hline
				total&
				1,005,000&
				\makecell[c]{53481\\(87.44)}&
				\makecell[c]{56686\\(92.68)}&
				\textbf{\makecell[c]{61164\\(100)}}\\ \hline 
			\end{tabular}
		\end{minipage}
	\end{center}
	\begin{center}
		{\bf (b) result on the rich dictionary}
		\begin{minipage}[t]{0.95\textwidth}
			\begin{tabular}{|c|c|c|c|c|} 
				\hline
				N-gram& 
				size& 
				sembei&
				PMI&
				PFNE \\ \hline
				1-gram&
				5,000&
				\makecell[c]{5000\\(100)}&
				\makecell[c]{5000\\(100)}&
				\textbf{\makecell[c]{5000\\(100)}}\\ \hline
				2-gram& 
				300,000& 
				\makecell[c]{49483\\(90.58)}&
				\makecell[c]{51452\\(94.18)}&
				\textbf{\makecell[c]{54632\\(100)}} \\ \hline
				3-gram& 
				300,000& 
				\makecell[c]{11221\\(62.53)}&
				\makecell[c]{15057\\(83.91)}&
				\textbf{\makecell[c]{17945\\(100)}}\\ \hline
				4-gram& 
				300,000& 
				\makecell[c]{11133\\(76.34)}&
				\makecell[c]{12963\\(88.89)}&
				\textbf{\makecell[c]{14584\\(100)}}\\ \hline
				5-gram& 
				50,000& 
				\makecell[c]{234\\(74.29)}&
				\makecell[c]{262\\(83.17)}&
				\textbf{\makecell[l]{315\\(100)}}\\ \hline
				6-gram& 
				50,000& 
				\makecell[c]{115\\(80.99)}&
				\textbf{\makecell[c]{142\\(110.07)}}&
				\makecell[c]{129\\(100)}\\ \hline
				total&
				1,005,000&
				\makecell[c]{77186\\(83.34)}&
				\makecell[c]{84876\\(91.65)}&
				\textbf{\makecell[c]{92605\\(100)}}\\ \hline
			\end{tabular}
		\end{minipage}
	\end{center}
\end{table}

As expected, our method greatly increases the number of valid n-gram in the embeding vocabulary and achieve the highest Precision and Recall comparing with SGNS-sembei and SGNS-PMI. In basic dictionary, the total number of valid n-gram of SGNS-sembei and SGNS-PMI is increased by 12.6\% and 7.4\%. In rich dictionary, the total number of valid n-grams of SGNS-sembei and SGNS-PMI is increased by 16.7\% and 8.4\%.

Furthermore, difference between the rich dictionary and the basic one is that the former contains much more neologism and informal words. Comparing with Table 2(a), the precentage of SGNS-sembei relative to SGNS-PFNE in Table 2(b) became less and this indicates that SGNS-PFNE is able to capture more neologisms and informal words from texts. Therefore, PFNE is more effective in the open domain situation, such as SNS data. The same phenomenon also lies in the comparision between SGNS-PMI and SGNS-PFNE.

\begin{figure}[!t]
	\caption{Precision-Recall curve for three methods.}
	\begin{minipage}{0.5\textwidth}
		\centering
		{\bf (a)Precision-Recall curve on the rich dictionary}
		\includegraphics[width=7cm]{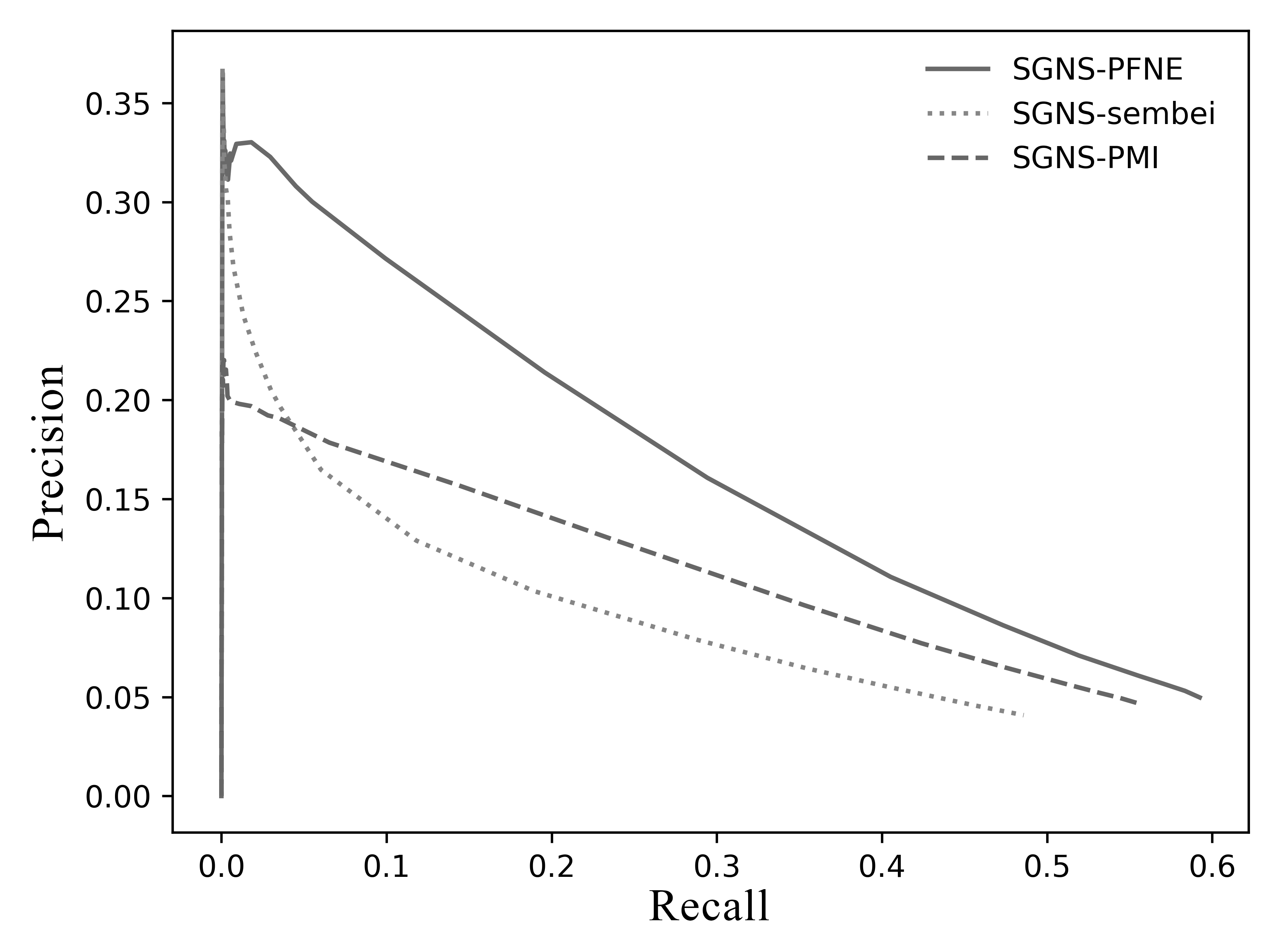}
		\centering
	\end{minipage}
	\begin{minipage}{0.5\textwidth}
		\centering
		{\bf (b)Precision-Recall curve on the basic dictionary}
		\includegraphics[width=7cm]{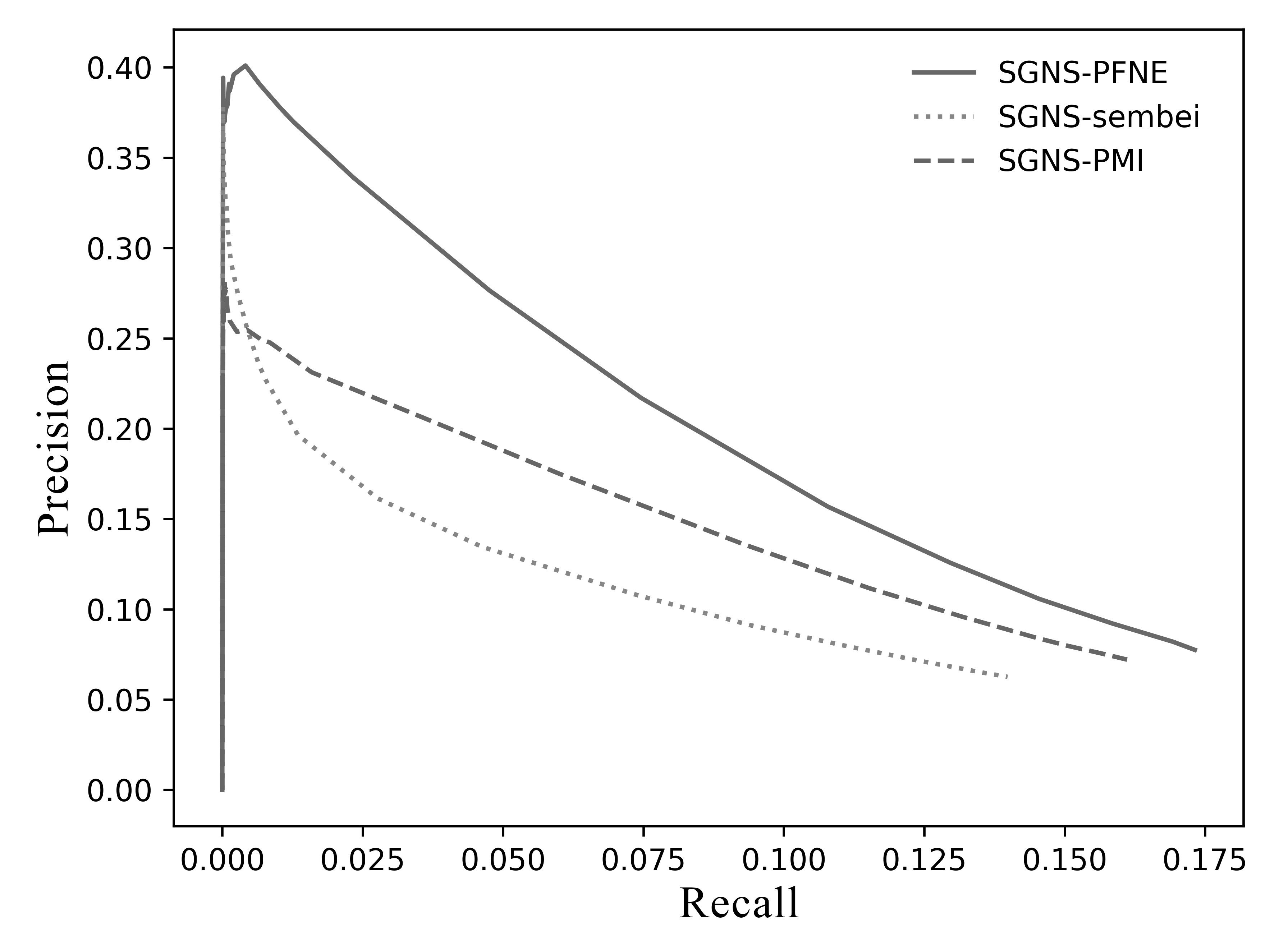}
	\end{minipage}
\end{figure}

\subsection{Results of experiments \uppercase\expandafter{\romannumeral2}}
The result of noun category prediction is shown in Table 3. It is observed that our method outperformes baselines in Precision, Recall and F1 scores. Comparing with the original segmentation-free word embedding model SGNS-sembei, the F-1 score is improved by 3.7\% with the proposed method.

We believe the reason why word embedding of SGNS-PFNE achieves better performance in the downstream task is that SGNS-PFNE can leverage more trustful statistical features from the corpus with a new unsupervised technique while SGNS-sembei only considers numerical raw occurrence frequency information. Specifically, SGNS-PFNE successfully reduces noise in the embedding targets by collecting more valid n-grams that are strongly associated in the contexts. And embedding vocabulary with higher quality can provide more words and theirs contexts with high association strength in the n-gram lattice which are contributed to word representation as training samples in the training phrase.

\begin{table}[]
	\caption{\bf Weighted Precision, Recall and F1 score of noun category prediction.}
	\setlength{\tabcolsep}{4mm}{
		\begin{tabular}{|c|c|c|c|}
			\hline
			Model& 
			Precision& 
			Recall&
			F1  \\ \hline
			Skip-Gram&
			0.672&
			0.682&
			0.669\\ \hline
			SGNS-sembei& 
			0.711& 
			0.706& 
			0.705\\ \hline
			SGNS-PMI& 
			0.726& 
			0.725&
			0.723\\ \hline
			SGNS-PFNE&
			\textbf{0.751}& 
			\textbf{0.733}& 
			\textbf{0.742}\\ \hline
		\end{tabular}
	}
\end{table}

\section{Conclusion and future work}

We proposed PFNE, which combines a new effective unsupervised association measure with distributed n-gram embedding model. Compared with the original segmentation-free word embedding models, the proposed model can filter out more associated valid n-gram(including informal words and neologisms) by utilizing more information from the corpus and thus construct a better n-gram lattice which is conducive to improve the performance of word vectors in the downstream task. We believe that our work can be particularly effective in dealing with unsegmented language problems, especially in the real-world situations, such as SNS data.

Chinese is a language rich in kinds of features. The future work to enhance segementation-free word embedding for Chinese can be divided into two directions. First way is to combine other word embedding techniques with PFNE, such as Glove$^{[7]}$ and BERT$^{[31]}$. Second approach can be using more information such as font structure and font pattern features to improve the capablity of word embeddings in capturing more latent relations from the corpus.

%----------------------------------------------------------------------------
%	REFERENCE LIST
%----------------------------------------------------------------------------

%\bibliography{ref}

\end{document}